\documentclass[sn-mathphys,Numbered,ePrint]{sn-jnl}
\usepackage[utf8]{inputenc}
\usepackage{graphicx}%
\usepackage{multirow}%
\usepackage{amsmath,amssymb,amsfonts}%
\usepackage{amsthm}%
\usepackage{mathrsfs}%
\usepackage[title]{appendix}%
\usepackage{xcolor}%
\usepackage{textcomp}%
\usepackage{manyfoot}%
\usepackage{booktabs}%
\usepackage{algorithm}%
\usepackage{algorithmicx}%
\usepackage{algpseudocode}%
\usepackage{listings}%
\usepackage{listingsutf8}
\usepackage{xcolor}

\definecolor{codegreen}{rgb}{0,0.6,0}
\definecolor{codegray}{rgb}{0.5,0.5,0.5}
\definecolor{codepurple}{rgb}{0.58,0,0.82}
\definecolor{backcolour}{rgb}{0.95,0.95,0.92}

\lstdefinestyle{python}{
    backgroundcolor=\color{backcolour},   
    commentstyle=\color{codegreen},
    keywordstyle=\color{magenta},
    numberstyle=\tiny\color{codegray},
    stringstyle=\color{codepurple},
    basicstyle=\ttfamily\footnotesize,
    breakatwhitespace=false,         
    breaklines=true,                 
    captionpos=b,                    
    keepspaces=true,                 
    numbers=left,                    
    numbersep=5pt,                  
    showspaces=false,                
    showstringspaces=false,
    showtabs=false,                  
    tabsize=2
}
\begin{document}

\title[Cabuxa Conversational Language Model]{Conversations in Galician: a Large Language Model for an Underrepresented Language}


\author*[1]{\fnm{Eliseo} \sur{Bao}}\email{eliseo.bao@udc.es}

\author[1]{\fnm{Anxo} \sur{Pérez}}\email{anxo.pvila@udc.es}

\author[1]{\fnm{Javier} \sur{Parapar}}\email{javier.parapar@udc.es}

\affil[1]{\orgdiv{Information Retrieval Lab, Centro de Investigación en Tecnoloxías da Información e da Comunicación (CITIC)}, \orgname{Universidade da Coruña}, \orgaddress{\street{Campus de Elviña s/n}, \city{A Coruña}, \postcode{15071}, \state{Galicia}, \country{Spain}}}


\abstract{
The recent proliferation of Large Conversation Language Models has highlighted the economic significance of widespread access to this type of AI technologies in the current information age. Nevertheless, prevailing models have primarily been trained on corpora consisting of documents written in popular languages. The dearth of such cutting-edge tools for low-resource languages further exacerbates their underrepresentation in the current economic landscape, thereby impacting their native speakers. This paper introduces two novel resources designed to enhance Natural Language Processing (NLP) for the Galician language. We present a Galician adaptation of the Alpaca dataset, comprising 52,000 instructions and demonstrations. This dataset proves invaluable for enhancing language models by fine-tuning them to more accurately adhere to provided instructions. Additionally, as a demonstration of the dataset utility, we fine-tuned LLaMA-7B to comprehend and respond in Galician, a language not originally supported by the model, by following the Alpaca format. This work contributes to the research on multilingual models tailored for low-resource settings, a crucial endeavor in ensuring the inclusion of all linguistic communities in the development of Large Language Models. Another noteworthy aspect of this research is the exploration of how knowledge of a closely related language, in this case, Portuguese, can assist in generating coherent text when training resources are scarce. Both the Galician Alpaca dataset and Cabuxa-7B are publicly accessible on our Huggingface Hub, and we have made the source code available to facilitate replication of this experiment and encourage further advancements for underrepresented languages.

}




\maketitle

\section{What is Cabuxa-7B?}\label{wn}

Cabuxa-7B\footnote{\url{https://huggingface.co/irlab-udc/cabuxa-7b}} is a LLaMA-7B \cite{touvron2023llama} LoRA \cite{hu2021lora} instruct-tuned model for Galician that can answer instructions in the Alpaca format\footnote{\url{https://github.com/tloen/alpaca-lora/blob/main/templates/alpaca.json}}. This work broadens the Portuguese effort from \citet{larcher2023cabrita} to Galician. Cabuxa-7B is intended to address a pressing need in the realm of low-resource languages, particularly for Galician. Low-resource languages often lack robust language models, making natural language processing tasks challenging in these linguistic contexts. 

LLaMA, which stands for Large Language Model Meta AI, is a family of large language models introduced by Meta AI in February 2023. These models come in various sizes, including 7 billion, 13 billion, 33 billion, and 65 billion parameters. 

Traditional fine-tuning of large language models for specific tasks can be prohibitively expensive in terms of computational resources. Low-Rank Adaptation of Large Language Models (LoRA) offers a novel approach to this problem. It involves preserving the pre-trained model's weights and introducing trainable rank decomposition matrices into each layer of the Transformer  architecture \cite{vaswani2023attention}. This approach significantly reduces the number of trainable parameters for downstream tasks. In comparison to conventional fine-tuning, LoRA can reduce the number of trainable parameters by a factor of 10,000 and reduce the GPU memory requirements by threefold.

\section{Data}

We translated the Alpaca \cite{alpaca} dataset to Galician with the Python package \texttt{googletranslatepy}\footnote{\url{https://suqingdong.github.io/googletranslatepy/}}. Cabuxa-7B was fed with the 80\% of this new dataset\footnote{\url{https://huggingface.co/datasets/irlab-udc/alpaca_data_galician}}, as we are keeping the remaining 20\% for future evaluation and experiments. 

\section{Training procedure}

We trained the model for 20 epochs with a Transformers \cite{wolf-etal-2020-transformers} Trainer object. This object was configured with the following \texttt{TrainingArguments}:
\begin{itemize}
    \item per\_device\_train\_batch\_size: 64
    \item gradient\_accumulation\_steps: 32
    \item warmup\_steps: 100
    \item num\_train\_epochs: 20
    \item learning\_rate: 3e-4
    \item fp16=True
\end{itemize}

LoRA and quantization \cite{dettmers2022optimizers} configurations are available both in the repository we release with this work\footnote{\url{https://gitlab.irlab.org/irlab/cabuxa}} and the Huggingface model card. Table \ref{tab1} shows the loss values for each training epoch.

\begin{table}[ht]
\
\caption{Training loss for each epoch}\label{tab1}%
\begin{tabular}{@{}ll@{}}
\toprule
Epoch & Loss \\
\midrule
        0.98  & 2.610 \\
        1.97  & 2.059 \\
        2.95  & 1.509 \\
        3.93  & 1.379 \\
        4.92  & 1.284 \\
        5.9   & 1.208 \\
        6.88  & 1.150 \\
        7.86  & 1.117 \\
        8.85  & 1.087 \\
        9.83  & 1.066 \\
        10.81 & 1.051 \\
        11.8  & 1.036 \\
        12.78 & 1.025 \\
        13.76 & 1.016 \\
        14.75 & 1.011 \\
        15.73 & 1.003 \\
        16.71 & 0.996 \\
        17.7  & 0.998 \\
        18.68 & 0.992 \\
        19.66 & 0.990 \\
\botrule
\end{tabular}
\end{table}

\section{Future steps}

Future work includes improving the translation of the dataset. It would also be desirable to be able to extend it to a wider variety of sources and tasks. Another important issue for the future is evaluation, which would ideally involve the use of expert linguists. Finally, we also intend to train and release versions of the model with a larger number of parameters.

\section{Environmental Impact}

The experiments were conducted using a private infrastructure. A cumulative of 72 hours of computation were performed on hardware of type NVIDIA RTX A6000. Total emissions are estimated to be 9.33 Kg. CO$_2$eq. Carbon emissions were estimated using the Machine Learning Impact calculator\footnote{\url{https://mlco2.github.io/impact/}} presented by \citet{lacoste2019quantifying}.

\newpage
\begin{appendices}

\section{How to get started with the model}\label{secA1}

Use the code below to get started with the model:

\begin{lstlisting}[language=Python, caption=Cabuxa 7-B playground example.]
from peft import PeftModel
from transformers import AutoModelForCausalLM, LlamaTokenizer, GenerationConfig

config = PeftConfig.from_pretrained("irlab-udc/cabuxa-7b")
model = AutoModelForCausalLM.from_pretrained("huggyllama/llama-7b", device_map="auto")
model = PeftModel.from_pretrained(model, "irlab-udc/cabuxa-7b")
tokenizer = LlamaTokenizer.from_pretrained("huggyllama/llama-7b")

# This function builds the prompt in Alpaca format
def generate_prompt(instruction, input=None):
    if input:
        return f"""Abaixo está unha instrución que describe unha tarefa, xunto cunha entrada que proporciona máis contexto. 
               Escribe unha resposta que responda adecuadamente a entrada.
               ### Instrución:
               {instruction}
               ### Entrada:
               {input}
               ### Resposta:"""
    else:
        return f"""Abaixo está unha instrución que describe unha tarefa.
               Escribe unha resposta que responda adecuadamente a entrada.
               ### Instrución:
               {instruction}
               ### Resposta:"""


def evaluate(instruction, input=None):
    prompt = generate_prompt(instruction, input)
    inputs = tokenizer(prompt, return_tensors="pt")
    input_ids = inputs["input_ids"].cuda()
    generation_output = model.generate(
        input_ids=input_ids,
        generation_config=GenerationConfig(do_sample=True),
        return_dict_in_generate=True,
        output_scores=True,
        max_new_tokens=256,
    )
    for s in generation_output.sequences:
        output = tokenizer.decode(s)
        print("Resposta:", output.split("### Resposta:")[1].strip())

evaluate("Cal é a fórmula química da auga?")
evaluate(
    "Convence ao lector por que é importante un determinado tema.",
    "Por que é esencial priorizar o sono?",
)

\end{lstlisting}




\end{appendices}


\newpage
\bibliography{sn-bibliography}

\end{document}